\documentclass[11pt,a4paper]{article}
\usepackage{authblk}
\usepackage{adjustbox}
\usepackage[hyperref]{acl2020}
\usepackage{times}  
\usepackage{latexsym}

\usepackage{adjustbox}
\usepackage{framed}
\usepackage{graphicx}
\graphicspath{{./}}
\usepackage{amsfonts} 
\usepackage{amsmath}
\usepackage[
singlelinecheck=false 
]{caption}
\usepackage{microtype}

\aclfinalcopy 

\title{Combining Context-Free and Contextualized Representations for Arabic Sarcasm Detection and Sentiment Identification}

\author[1]{\textbf{Amey Hengle}}
\author[1]{\textbf{Atharva Kshirsagar}}
\author[1]{\textbf{Shaily Desai}}
\author[2]{\textbf{Manisha Marathe}}

\affil[1,2]{Department of Computer Engineering, PVG's College of Engineering and Technology,\protect \\ affiliated to Savitribai Phule Pune University, India.}

\affil[1]{\texttt {{\{ameyhengle22, atharvakshirsagar145, shailysd02\}} @gmail.com}}
\affil[2]{\texttt {mvm\_comp@pvgcoet.ac.in }}
\date{} 

\begin{document}
\maketitle
\begin{abstract}

Since their inception, transformer-based language models have led to impressive performance gains across multiple natural language processing tasks. For Arabic, the current state-of-the-art results on most datasets are achieved by the AraBERT language model. Notwithstanding these recent advancements, sarcasm and sentiment detection persist to be challenging tasks in Arabic, given the language's rich morphology, linguistic disparity and dialectal variations. This paper proffers team SPPU-AASM's submission for the WANLP ArSarcasm shared-task 2021, which centers around the sarcasm and sentiment polarity detection of Arabic tweets. The study proposes a hybrid model, combining sentence representations from AraBERT with static word vectors trained on Arabic social media corpora. The proposed system achieves a F1-sarcastic score of 0.62 and a F-PN score of 0.715 for the sarcasm and sentiment detection tasks, respectively. Simulation results show that the proposed system outperforms multiple existing approaches for both the tasks, suggesting that the amalgamation of context-free and context-dependent text representations can help capture complementary facets of word meaning in Arabic. The system ranked second and tenth in the respective sub-tasks of sarcasm detection and sentiment identification.
\end{abstract}

\section{Introduction}
With the advent of social media platforms as a valuable source of opinion-rich information, work on subjectivity language analysis has continued to receive increasing interest from the Arabic NLP community. Sentiment Analysis (SA) has been the dominant theme in this area, with notable works ranging from the creation of lexical resources and sentiment datasets \citep{el-beltagy-2016-nileulex, badaro-etal-2014-large, Elmadany2018ArSASA, kiritchenko-etal-2016-semeval} to the contrivance of neural network-based classification models \citep{Alayba_2018,Heikal2018SentimentAO, Kaibi2020SentimentAA}. In comparison, the literature in Arabic sarcasm detection is still in its nascent stage, limited to a few notable works \citep{karoui2017soukhria, ghanem2019idat, abbes2020daict}.

Recently, transformer-based language models have proved highly efficient at language understanding, achieving promising results across multiple NLP tasks and benchmark datasets.  The language modeling capability of these models is aiding in capturing the literal meaning of context-heavy texts. For Arabic NLP in particular, the best results for sentiment analysis are currently achieved by AraBERT, a language model proposed by \citet{antoun2020arabert}. 

Despite this recent progress, sarcasm detection remains a challenging task, primarily due to the use of implicit, indirect phrasing and the figurative nature of language \citep{abufarha-etal-2021-arsarcasm-v2}. The task becomes even more challenging when working with Twitter data, as the social media posts tend to be short and often contain sources of noise, code-switching, and the use of nonstandard dialectal variations \citep{baly-etal-2017-characterization}. Furthermore, BERT-based models are found to struggle with rare words \citep{schick2019rare}, which can be encountered more in social media texts due to their informal nature and the prevalent use of slang words. For language models like AraBERT, this can pose a challenge, primarily since it has been trained on structured corpora from Wikipedia. 

Building on the capabilities of language models, some recent studies have shown that incorporating entity vectors can benefit the BERT-based language models, especially for domain-specific tasks or datasets  \citep{poerner-etal-2020-e, lin2019kagnet,peinelt-etal-2020-tbert}. An interesting approach followed by \citet{alghanmi-etal-2020-combining}  suggests that the performance of language models can be boosted by incorporating static word embeddings trained on specific social media corpora. 

In this study, we posit a solution for subtask-1 and subtask-2 of the WANLP ArSarcasm shared task 2021 \citep{abufarha-etal-2021-arsarcasm-v2}. While subtask-1 focuses on identifying sarcasm, subtask-2 deals with classifying the sentiment polarity in Arabic tweets. Inspired by the works of \citet{peinelt-etal-2020-tbert} and \citet{alghanmi-etal-2020-combining}, we propose a hybrid model, combining the sentence representations learned from AraBERT with pre-trained Arabic word vectors proposed by \citet{abu-farha-magdy-2019-mazajak}. Results attest that the proposed methodology can provide a competent way of subsuming the advantages of both the contextualized and context-free word representations, outperforming all the baseline models for both tasks.

The rest of the paper is organized as follows: In Section ~\ref{section2}, we provide a concise literature review of previous works in Arabic sentiment and sarcasm detection. Section ~\ref{section3}  provides a descriptive analysis of the dataset at hand. In Section ~\ref{section4}, we present the proposed system, and in Section ~\ref{section5}, we describe the experimental setup details. Section ~\ref{section6} interprets the results. Finally, Section ~\ref{section7} concludes the study and points to possible directions for future work.

\section{Related Work}
\label{section2}
Early works in Arabic subjectivity analysis focused on using conventional machine learning approaches and lexical methods \citep{al2019comprehensive}.  With the emergence of deep learning techniques, research in Arabic NLP has shifted from the traditional statistical standpoint to designing complex neural network models and learning word representations.  \citet{al-sallab-etal-2015-deep} experimented with various deep learning models such as a recursive autoencoder (RAE), deep belief networks (DBN), and a deep auto-encoder (DAE). \citet{Alayba_2018} built an Arabic SA system based on a combination of CNNs and LSTMs.  In \citep{AlSmadi2018DeepRN}, the authors proposed an aspect-based sentiment analysis system based on a hybrid architecture of BiLSTM and conditional random field (CRF).

The success of the English word2vec \citep{mikolov2013efficient} and fast-Text \citep{bojanowski2016enriching} motivated other works to achieve the same feat by creating language-specific word embeddings. For Arabic NLP, some early attempts include word2vec-based AraVec \citep{SOLIMAN2017256}, followed by fast-Text \citep{bojanowski2016enriching}. Recently, \citet{abu-farha-magdy-2019-mazajak} proposed the Mazajak embeddings, trained exclusively on a large Arabic twitter corpus for handling the varied Arabic dialects. Multiple studies leveraged this advancement in word representations for both the sarcasm and sentiment detection tasks. For instance, \citet{Heikal2018SentimentAO} developed a CNN and LSTM ensemble model for Arabic SA. The authors employed pre-trained AraVec word embeddings for the text representation. \citet{Kaibi2020SentimentAA} proposed a hybrid model for Arabic SA, concatenating pre-trained AraVec and fast-Text vectors. \citet{abu-farha-magdy-2019-mazajak}  used the pre-trained Mazajak vectors on a CNN-BiLSTM ensemble model, achieving state-of-the-art results on three benchmark datasets. For sarcasm detection, a similar approach was followed by \citet{abufarha-etal-2021-arsarcasm-v2}, where the authors used the Mazajak vectors in combination with a BiLSTM model.

The best results for multiple datasets is currently achieved by fine-tuning the AraBERT model \citep{antoun2020arabert}, eliminating the need to use the static word vectors in standard settings. Despite this fact, we believe that AraBERT and Mazajak have complementary strengths and can lead to improved results if used in coalescence. In this study, we investigate the effectiveness of combining the word representations obtained from these two models on the sarcasm detection and sentiment identification tasks.

\section{Dataset}
\label{section3}
The WANLP ArSarcasm shared-task 2021 follows the ArSarcasm v2 dataset \citep{abufarha-etal-2021-arsarcasm-v2}. The dataset contains sarcasm, sentiment and dialect labels of 12,549 Arabic tweets. The tweets span across five Arabic dialects including MSA, Gulf, Egyptian, Levantine, and Maghrebi, with MSA and Egyptian dialects having the highest percentage of tweets. 
For development, we follow a standard 80-20 stratified split on the ArSarcasm v2 dataset. This leaves us with a validation set of 2510 tweets, which are used for the primary evaluation of the proposed system along with the baseline models.  The organizers provide a separate dataset for testing, consisting of 3,000 tweets. Table ~\ref{table1} and Table ~\ref{table2} provide a descriptive analysis of the final training, validation and test sets for the tasks of sentiment identification and sarcasm detection respectively.

\begin{table}[htp]
\centering
\begin{adjustbox}{width=\columnwidth}
\resizebox{\columnwidth}{!}{%
\begin{tabular}{ccccc}
\hline \textbf{Set} & \textbf{Positive} & \textbf{Negative} & \textbf{Neutral} & \textbf{Total}\\
\hline \hline
Training & 1744 & 3697 & 4598 & 10039\\
Validation & 436 & 925 & 1149 & 2510\\
Testing & 575 & 1677 & 748 & 3000\\
\hline\hline
{Total} & 2755 & 6298 & 6495 & 15548\\ 
\hline
\end{tabular}%
}
\end{adjustbox}
\caption{\label{table1}Label wise distribution for sentiment analysis. }
\end{table}

\begin{table}[htp]
\label{table 2}
\centering
\begin{tabular}{cccc}
\hline 
\textbf{Set} & \textbf{True} & \textbf{False} & \textbf{Total}\\
\hline \hline
Training & 1734 & 8305 & 10039\\
Validation & 434 & 2076 & 2510\\
Testing & 821 & 2179 & 3000\\
\hline\hline
{Total} & 2989 & 12559 & 15548\\ 
\hline
\end{tabular}
\caption{\label{table2}Label wise distribution for sarcasm detection. }
\end{table}

\section{Proposed System}
\label{section4}
\begin{figure}[ht]

    \includegraphics[width = \linewidth, height= 10 cm]{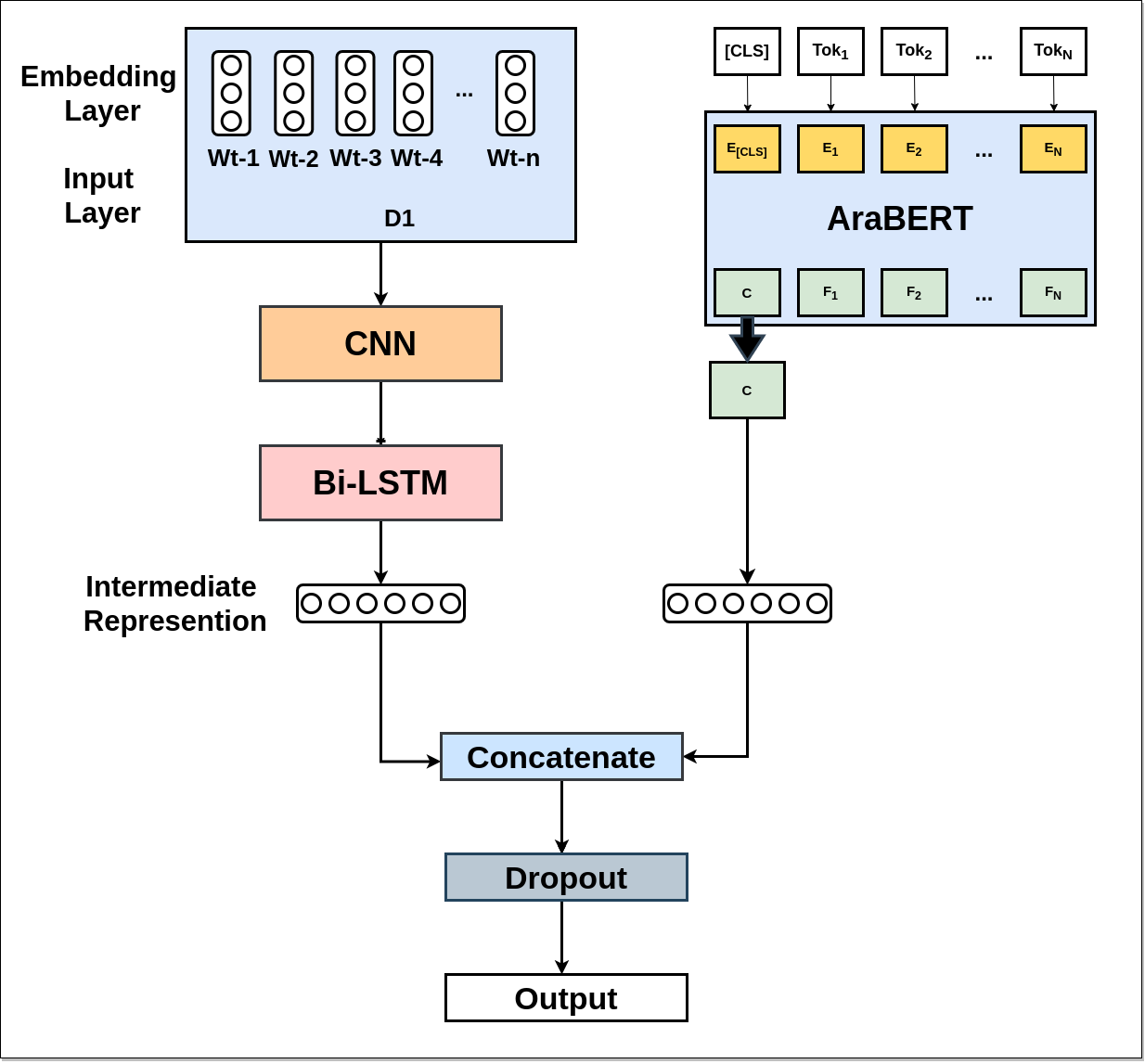}
    
\caption{\label{fig1} Architecture diagram of the proposed model.}
\end{figure}

Unlike the static word models such as word2vec or fast-Text, language models like AraBERT follow a different tokenization strategy; wherein each word is split into one or more wordpiece tokens \citep{antoun2020arabert}. Thus, we cannot simply concatenate the pre-trained Mazajak embeddings with the contextualized representations predicted by AraBERT at the word level. We instead combine these word representations at the sentence level, following an approach similar to the one used by \citet{peinelt-etal-2020-tbert}. The final sentence representation is passed to the dropout layer, followed by a dense layer with Softmax activation for classification.  Fig ~\ref{fig1}  gives an overview of the proposed system’s architecture. The following section describes each system component in detail.    

\subsection {CNN-BiLSTM emsemble}
In order to get a sentence representation from the static word embeddings, we employ a CNN-BiLSTM ensemble model. The model learns a 128-dimensional feature vector from the pre-trained Mazajak embeddings. While CNN excels at extracting features from the input data, BiLSTM supports better modeling of sequential dependencies. Thus, using an ensemble helps us subsume the advantages of both these techniques.

\subsubsection{Static Embedding Input}
The proposed system makes use of the Skip-Gram model variant of the Mazajak word embeddings, pre-trained on 250 million Arabic tweets \footnote{http://mazajak.inf.ed.ac.uk:8000/} . During training, the embedding layer maps each word in a tweet to its corresponding embedding, converting the tweet into a real valued dense vector.

\subsubsection{CNN Layer}
The word vectors generated by the embedding layer are fed to a CNN layer. The discrete convolutions performed by the CNN layer help to extract the most influential n-grams in the tweet. This is followed by a max-pooling layer. 

\subsubsection{BiLSTM Layer}
The bidirectional short term memory (BiLSTM) layer is just a combination of two LSTMs running in opposite direction \citep{graves2005framewise}, allowing the network to simultaneously  encode the forward and backward information of a sequence at each time step. The intermediate sentence representation generated by the CNN layer is passed to the BiLSTM layer, which encodes it into a 128-dimensional feature vector $D  \in \mathbb{R}^{d_1}$. 

\subsection{AraBERT}
The contextualized tweet representations are obtained using the pre-trained  AraBERT model \citep{antoun2020arabert}. In particular, each tweet $t$ is encoded using the $C$ vector from AraBERT’s final hidden state layer corresponding to the special classification token [CLS]:

\begin{equation}
\label{cls-token}
C = AraBERT (t) \in \mathbb {R}^{d_2}
\end{equation}

where $d_2$ denotes the internal hidden size of AraBERT (768 for AraBERT v0.2). 

Computed using self-attention, the [CLS] token vector is designed to collect information from the rest of the hidden states and be used as a unique representation of the entire sequence \citep{devlin2019bert}. We find that this method is more robust than averaging the hidden states, mainly since it avoids every state to be averaged with the same weight, including stopwords and tokens not relevant to the classification task.

\subsection{Combined Classification Layer}
The final sentence representation $F$ is obtained by simply concatenating the $C$ vector predicted by the AraBERT model with the feature vector $D$ obtained from the CNN-BiLSTM ensemble. 

\begin{equation}
\label{concat}
F = [D; C]  \in \mathbb {R}^{d1 + d2 }
\end{equation}

After applying a dropout, the resultant concatenated vector is passed to a dense layer with Softmax activation for classification.

\section{Experimental Setup}
\label{section5}
\subsection{Data Preparation}
For each tweet in the corpus, we apply standard text cleaning steps including the removal of hashtags, mentions, urls, punctuations and arabic diacritics \citep{said2009study}. For stopword removal, we used a publicly available resource \footnote{https://github.com/mohataher/arabic-stop-words}. To replace emojis and emoticons with their corresponding Arabic translations, we created a custom dictionary mapping. Arabic text normalization and lemmatization is done using the AraBERT preprocessor \footnote{https://github.com/aub-mind/arabert}. Each tweet is padded to a maximum length of 100 for both the AraBERT and the CNN-BiLSTM model. Longer tweets are truncated.

\subsection{Parameters and Training Environment}
For the CNN-BiLSTM ensemble part of the proposed system, we employ a CNN layer with 256 filters and relu activation. For the BiLSTM layer, we use 128 dimensional units (64 for each LSTM), and apply a recurrent dropout rate of 0.2. All layers are implemented using Keras \footnote{https://keras.io/about/}.
For using the pre-trained AraBERT model, we follow the Tensorflow implementation of Hugging Face \footnote{https://github.com/huggingface/transformers}. We make use of the AraBERT v0.2 version, which is officially available under the name bert-base-arabertv02. The same model is used for text tokenization. After predicting the [CLS] token vector from the AraBERT model, we freeze the model weights. We then jointly train the CNN-BiLSTM ensemble with the combined classification layer. The model is trained using the Adam optimizer \citep{kingma2017adam}, a learning rate of 5e-5, epsilon value of 1e-08, clipnorm equal to 1.0, a batch size of 16, and sparse categorical crossentropy loss, with the usage of early stopping for a callback.

\subsection{Baseline Models}
As a baseline, we compare the performance of the proposed system against the results of the pretrained AraBERT model  \citep{antoun2020arabert}, for both the sarcasm and sentiment detection tasks. Furthermore, we show the performance of the previous systems proposed by  \citet{abu-farha-magdy-2020-arabic} and \citet{abu-farha-magdy-2019-mazajak} for the tasks of sarcasm detection and sentiment identification respectively.

\begin{table*}[!htb]
\captionsetup{justification=centering,margin=1cm}
\centering
\begin 
{tabular}{llllll}
\\\hline 
\textbf{Model} & \textbf{Accuracy} & \textbf{Precision} & \textbf{Recall} & \textbf{F1-} & \textbf{F1-}\\ & & & & \textbf{Macro}& \textbf{Sarcastic} \\ \hline \hline
\citep{abu-farha-magdy-2020-arabic} & - & 0.62 & 0.38 & 0.46 & 0.44\\
AraBERT & 0.85 & 0.75 & 0.70 & 0.72 & 0.52\\
    \textbf{AraBERT + CNN-BiLSTM} & \textbf{0.86} & \textbf{0.76}
& \textbf{0.78} & \textbf{0.77} & \textbf{0.62}\\ 

\hline \\
\textbf{AraBERT + CNN-BiLSTM} \\ {(Official results on test set)} & \textbf{0.7410} & \textbf{0.7031}
& \textbf{0.7447} & \textbf{0.7096} & \textbf{0.6140}\\
\hline \hline

\end{tabular}
\caption{\label{table3} Performance comparison of models for subtask-1 : sarcasm detection. All metrics correspond to the results on the sarcastic class.}

\begin{tabular}{llllll}
\\\hline \textbf{Model} & \textbf{Accuracy} & \textbf{Precision} & \textbf{Recall} & \textbf{F1-Macro} & \textbf{F-PN} \\ \hline \hline
\citep{abu-farha-magdy-2019-mazajak} & 0.67 & 0.64 & 0.66 & 0.64 & 0.60\\
AraBERT & 0.73 & 0.71 & 0.68 & 0.70 & 0.67\\
\textbf{AraBERT + CNN-BiLSTM} & \textbf{0.75} & \textbf{0.72}
& \textbf{0.73} & \textbf{0.72} & \textbf{0.71}\\
\hline \\
\textbf{AraBERT + CNN-BiLSTM} \\ {(Official results on test set)} & \textbf{0.6840} & \textbf{0.6421}
& \textbf{0.6388} & \textbf{0.6232} & \textbf{0.7073}\\
\hline \hline
\end{tabular}
\caption{\label{table4} Performance
comparison of models for subtask-2: sentiment identification.}

\end{table*}

\section{Results and Discussion}
\label{section6}
The official evaluation metric for the sarcasm detection subtask is the F-score of the sarcastic class, while that for the sentiment identification subtask is the F-PN score (macro average of the F-score of the positive and negative classes). For each model, the validation set results are averaged over five runs to ensure a fair comparison. 

Table ~\ref{table3} shows the results for the sarcasm detection subtask. The proposed model shows a 10-percent improvement in the F1-sarcasm score over the baseline AraBERT model in identifying sarcastic tweets. This indicates that the proposed system offers a more nuanced ability to capture the figurative meaning of tweets and identify implicit negative sentiments. Table ~\ref{table4} lists the baseline models' performances and the proposed system for sentiment identification subtask.  It is observed that both the proposed and AraBERT baseline model perform well on the neutral class. However, a better F-PN score indicates that the proposed model can more efficiently distinguish the positive and negative sentiment polarities from the neutral class. Furthermore, the proposed system seems to better handle the data-imbalance for both the tasks and is more robust to overfitting on the minority classes, showing a significant lead in the recall scores.
\\\\
Overall, the proposed method shows improved results across all the metrics for both the sarcasm and sentiment detection tasks. The performance improvements can be attributed to the fact that unlike the Mazajak word embeddings, which are exclusively trained on a Twitter corpus, the AraBERT model is trained on the Arabic Wikipedia and news corpora, preventing it from witnessing the varied dialects in which social media posts are written. It is reasonable to postulate that while language models like AraBERT capture rich contextual information, the Mazajak word vectors can provide valuable complementary information for rare words, which can be found abundantly in social media texts. Hence, using them in combination can only help a system capture complementary facets of word meaning, thereby enhancing its performance on the downstream sentiment and sarcasm detection tasks. 
 
\section{Conclusion}
\label{section7}
In this study, we proposed a hybrid model to combine the contextualized sentence representations obtained from AraBERT with pre-trained Mazajak word vectors. We show that the proposed model outperforms the standalone AraBERT model for both the sarcasm and sentiment detection tasks. Our findings suggest that incorporating static word vectors might help language models like AraBERT to deal with rare words and the constantly updating language of social media. An alternative strategy would be to pre-train AraBERT on specific social media corpora like Twitter. However this can prove to be extremely expensive and is not feasible in practice. It is also important to note that, while the proposed model leads in performance, it is more complex, and has a greater number of trainable parameters. Hence, it would  be essential to test its feasibility on datasets larger than the ArSarcasm v2 dataset. 

\bibliography{asi_references}
\bibliographystyle{acl_natbib}

\end{document}